\documentclass[runningheads]{llncs}
\usepackage[T1]{fontenc}
\usepackage{booktabs,multirow}
\usepackage{tikz}
\usepackage{subcaption}
\usepackage{graphicx,verbatim}
\usepackage{algorithm}
\usepackage{algpseudocode}
\usepackage{amsfonts}
\usepackage{amsmath}
\usepackage{bm}

\usepackage{makecell}
\usepackage{pifont}
\newcommand{\cmark}{\ding{51}} % check mark
\newcommand{\xmark}{\ding{55}} % cross mark
\usepackage{array}
\usepackage[hyphens]{url}
\usepackage{hyperref}
\usepackage{multirow}  % for \multirow
\usepackage{booktabs}  % for \toprule, \midrule, \bottomrule

\algrenewcommand\algorithmicrequire{\textbf{Input:}}
\algrenewcommand\algorithmicensure{\textbf{Output:}}
\makeatletter
\renewcommand{\ALG@beginalgorithmic}{\small}
\makeatother
\floatname{algorithm}{Algorithm}

\begin{document}
\title{DBT-Bleed: Dual-Branch Temporal Modeling with Key-Frame Selection for Surgical Bleeding Detection}
\begin{comment}  %% Removed for anonymized MICCAI submission
\author{Sudhanshu Mishra\inst{1}\orcidID{0000-1111-2222-3333} \and
Jialang Xu\inst{2}\orcidID{1111-2222-3333-4444} \and Jensen Ang\inst{1}\orcidID{0000-1111-2222-3333} \and Evans B.\inst{1}\orcidID{0000-1111-2222-3333} \and Being-Ti Ang\inst{1}\orcidID{0000-1111-2222-3333} \and
Yueming Jin\inst{3}\orcidID{2222--3333-4444-5555}}
%
\authorrunning{Mishra et al.}
\institute{Princeton University, Princeton NJ 08544, USA \and
Springer Heidelberg, Tiergartenstr. 17, 69121 Heidelberg, Germany
\email{lncs@springer.com}\\
\url{http://www.springer.com/gp/computer-science/lncs} \and
ABC Institute, Rupert-Karls-University Heidelberg, Heidelberg, Germany\\
\email{\{abc,lncs\}@uni-heidelberg.de}}

\end{comment}

% \author{Aninymized}  %% Added for anonymized MICCAI submission
% \authorrunning{Anonymized Author et al.}
% \institute{Anonymized Affiliations \\
%     \email{email@anonymized.com}}

\author{Sudhanshu Mishra\inst{1}\thanks{Equal contribution} \and Jialang Xu\inst{3}\protect\footnotemark[1]\and Jensen Ang\inst{4} \and Evangelos B. Mazomenos\inst{3} \and Beng Ti Ang\inst{4} \and Yueming Jin\inst{1,2}\thanks{Corresponding author}}  %% Added for anonymized MICCAI submission
% index{Mishra, Sudhanshu}
% index{Xu, Jialang}
% index{Ang, Jensen}
% index{Mazomenos, Evangelos B.}
% index{Ang, Beng Tir}
% index{Jin, Yueming}

\authorrunning{Mishra et al.}
\institute{Department of Electrical and Computer Engineering, National University of Singapore, Singapore
% \\ \email{e1504738@u.nus.edu, ymjin@nus.edu.sg}
\and Department of Biomedical Engineering, National University of Singapore, Singapore \and  UCL Hawkes Institute, Department of Medical Physics and Biomedical Engineering, University College London, London, United Kingdom \and Department of Neurosurgery, National Neuroscience Institute, Singapore  \\
    \email{ymjin@nus.edu.sg}}
    %  add Hawkes institutes
  
\maketitle              % typeset the header of the contribution
\begin{abstract}
Intraoperative Adverse Events (IAEs) detection is critical for improving surgical safety, with bleeding being among the most frequent events across many surgery types. Existing methods struggle to distinguish bleeding IAE from visually similar residual blood due to limited temporal reasoning. Moreover, modeling long surgical videos while preserving fine-grained temporal dynamics remains computationally challenging. We propose \textit{DBT-Bleed}, a dual-branch multi-scale temporal modeling framework disentangling bleeding and normal representations using layer-wise temporal adapters for short- and long-term bleeding progression. To efficiently process long surgical videos without sacrificing fine-grained temporal information, we introduce HiRED, a Hierarchical Entropy-Driven frame selection strategy that retains temporally informative segments while removing redundancy. Experiments on the MultiBypass dataset demonstrate gains of $6.53\%$ in F1, $5.62\%$ in Recall and $9\%$ in MCC values for bleeding IAE detection, consistently outperforming video-level baselines.
Additionally, we evaluate cross-procedure generalization on a newly curated dataset from a different surgical procedure type, where DBT-Bleed demonstrates robust transferability by achieving gain of $6\%$ in F1 and $8\%$ in MCC under zero-shot setting. To support this evaluation, we introduce \textit{EndoPit-IAE}, an Endonasal Pituitary Surgery dataset annotated for IAEs, representing the first IAE-annotated dataset in neurosurgery. Code is available at: \href{https://github.com/jinlab-imvr/DBT-Bleed}{https://github.com/jinlab-imvr/DBT-Bleed}

% Authors must provide keywords and are not allowed to remove this Keyword section.

\end{abstract}
\section{Introduction}
Intraoperative Adverse Events (IAEs) are rare but critical occurrences during surgery. They can lead to severe postoperative complications, including infection, organ dysfunction, or even mortality~\cite{zegers2011incidence}, and often disrupt surgical workflow by necessitating immediate corrective interventions. They not only jeopardize patient safety but also increase healthcare costs due to prolonged recovery times. IAEs contribute upto 400k deaths annually~\cite{makary2016medical}, with bleeding being the most common at 23\%~\cite{zegers2011incidence}.

Therefore, it is crucial to timely detect bleeding to enable prompt intervention and improve surgical outcomes.

\begin{figure}[t]
  \centering

  \includegraphics[width=0.7\linewidth]{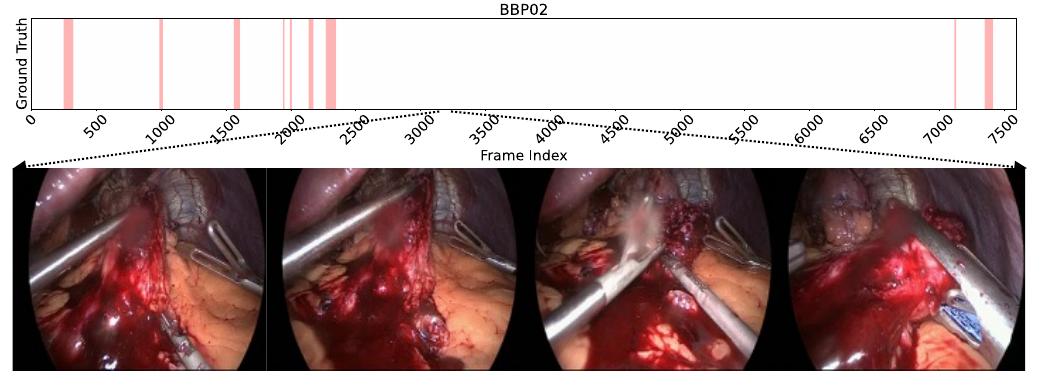}
  \caption{Non-IAE residual blood visibly similar to bleeding IAEs.}
  \label{mby140_weakness}
\end{figure}

Despite promising efforts~\cite{wei2021intraoperative}, video-level IAEs detection remains relatively underdeveloped. Methods focus on bleeding segmentation~\cite{miao2024hemoset,rabbani2022video,zou2025surgical}, blood source and flow detection~\cite{richter2021autonomous,hua2022automatic,pei2025synergistic,rahbar2020entropy,sogabe2023bleeding,xu2025bleedorigin}. However, detecting bleeding as an IAE differs from the tasks described above, because it requires distinguishing bleeding caused by a manual error from anticipated surgical bleeding.

The lack of IAEs-annotated public datasets has constrained research in this field, making the recent extension of MultiBypass dataset~\cite{lavanchy2024challenges,lavanchy2023proposal,ramesh2021multi,ramesh2023weakly} for IAEs annotation~\cite{bose2025feature,lavanchy2025analyzing} a vital advancement.
% The lack of publicly accessible datasets with detailed IAEs annotations has constrained systematic investigation in surgical adverse events detection.  The recent extension of the publicly available MultiBypass dataset with IAEs annotations~\cite{bose2025feature,lavanchy2025analyzing} constitutes a significant advancement for surgical IAEs detection.
The dataset naturally contains frames with visible blood that are labeled as ``normal'' (Fig.~\ref{mby140_weakness}), as the blood may originate from clinically resolved events or intentional surgical preparation. Such bleeding constitutes `Residual-bleeding'  leading to ambiguous frame-level supervision, whereas `Bleeding-IAE' is unintended surgical-site bleeding from mechanical or thermal incision.
% During MultiBypass dataset inspection, we found frames with visible blood labeled as “non-bleeding” because the blood originated from a clinically resolved event or intentional preparation and had not yet been cleared from the surgical field (Fig.~\ref{mby140_weakness}) leading to ambiguous frame-level supervision.
Horita et al.~\cite{horita2024real} applied YOLOv7 for frame-wise real-time bleeding detection without explicit temporal modeling. Bose et al.~\cite{bose2025feature} incorporated short-range temporal context but focused on limited clip lengths, restricting multi-scale modeling of bleeding progression. Consequently, two key challenges remain: (1) the need for robust temporal modeling to distinguish bleeding IAE from visually similar residual blood, and (2) the difficulty of efficiently modeling long surgical video sequences without losing fine-grained temporal dynamics.

Subsequently, (1) We propose DBT-Bleed, a dual-branch multi-scale temporal framework for surgical bleeding detection. A layer-wise Multi-scale Temporal Adapter(MTA) captures both short- and long-term bleeding progression to distinguish bleeding adverse events from residual blood. (2) We introduce  Hierarchical Entropy-Driven (HiRED) frame selection to process long surgical videos without sacrificing fine-grained dynamics. This strategy removes redundant segments while preserving temporally informative frames. (3) We demonstrate cross-procedure robustness of our method via zero-shot evaluation on a newly curated inhouse Endonasal Pituitary Surgery dataset. In this context, we introduce EndoPit-IAE, the first neurosurgical dataset annotated for IAEs. (4) We demonstrate state-of-the-art performance on both MultiBypass and EndoPit-IAE through extensive experiments and ablation studies.

\section{Methods}
\subsection{Dataset}
\textbf{MultiBypass}~\cite{bose2025feature,lavanchy2025analyzing} is a multi-center dataset of laparoscopic bypass surgery videos comprising 140 full-length surgical procedures, with corresponding frame-level IAE annotations at 1 fps available at~\cite{bose2025feature}. All training, validation, and testing experiments are conducted on MultiBypass. 

To further assess generalization, we curate \textbf{EndoPit-IAE}, an in-house Endonasal Pituitary Surgery dataset with detailed IAEs annotations, representing---to the best of our knowledge---the first video dataset dedicated to neurosurgical IAEs detection. Annotations were performed at 1-second intervals by expert surgeons. Three annotators labeled the in-house dataset: one consultant with > 10 years of experience and two consultant-trained trainees. Trainee annotations were reviewed by the consultant, and discrepancies were resolved by joint re-review and consensus, with an average inter-rater discrepancy < 0.5\% per video. Annotators estimated the flow rate of any bleeding present on the screen, and scored it using the VIBe (Validated Intraoperative Bleeding)~\cite{sciubba2022vibe} scale.
EndoPit-IAE serves exclusively as an external zero-shot benchmark evaluating cross-procedure robustness. Frame-level supervision is unreliable due to persistent residual blood and temporally sparse events in long videos (Fig. 1), motivating video-level supervision by aggregating $N=300$ frames with $100$ frame overlap into videos (Table.~\ref{tab:fold1_frames_to_clips}).

\begin{table*}[ht]
\centering
\caption{Dataset statistics}
\label{tab:fold1_frames_to_clips}
\begin{tabular}{l|lcc}
\toprule
\textbf{Dataset} & \textbf{Split} & \textbf{Non-bleed} & \textbf{Bleeding} \\
\midrule
\multirow{3}{*}{\textbf{MultiBypass}}
& Train & 1,586 & 600 \\
& Val   & 416   & 173 \\
& Test  & 884   & 267 \\
\midrule
\textbf{EndoPit-IAE}
& Test  & 64    & 128 \\
\bottomrule
\end{tabular}
\end{table*}

\subsection{Problem Definition}
Let $\mathcal{D}=\{(V_i, y_i)\}_{i=1}^{M}$ denote a set of surgical videos, where each video $V_i=\{x_{i,t}\}_{t=1}^{N_i}$ contains total $N_i$ RGB frames and is associated with a binary video-level label $y_i\in\{0,1\}$ indicating the absence or presence of bleeding IAE.
% We aim to learn a classifier $f_\theta$ that predicts a video-level probability $\hat{y}_i = f_\theta(V_i) \in [0,1]$ under weak supervision, labeling videos positive if a bleeding IAE occurs within their N frames.
We aim to learn a weakly supervised classifier $f_\theta$ that predicts a video-level probability $\hat{y}_i = f_\theta(V_i)$, with positive labels indicating at least one bleeding IAE in $V_i$.

\subsection{DBT-Bleed}
\subsubsection{Overview Architecture}
\begin{figure}[t]
    \centering
    \includegraphics[width=\textwidth]{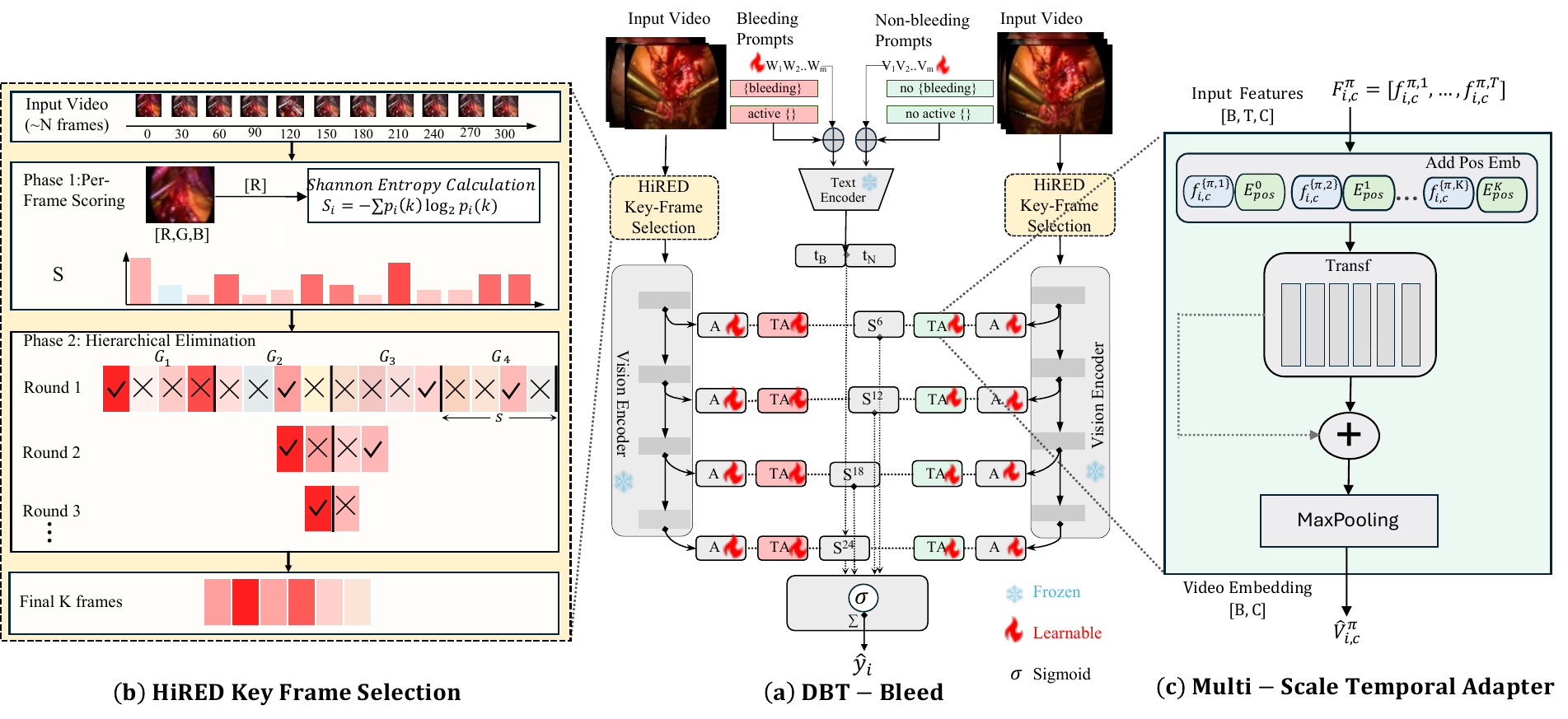}
    \caption{Overview of the proposed (a) DBT-Bleed framework, including (b) HiRED key frame selection, and (c) dual-branch multi-scale temporal adapter.}
    \label{fig:method_diagram}
\end{figure}

% We design DBT-Bleed, a CLIP-based framework with multi-scale temporal modeling, as shown in Fig.~\ref{fig:method_diagram}, to effectively capture the dynamic characteristics of adverse events in complex surgical scenes.
We propose DBT-Bleed---a CLIP-based dual-branch framework with multi-scale temporal modeling, shown in Fig.~\ref{fig:method_diagram}, to effectively capture the dynamic characteristics of adverse events in complex surgical scenes.
Given a long surgical video of $N$ frames, HiRED selects the top $K$ informative frames via Shannon entropy estimation followed by hierarchical pruning. The selected frames are processed by a frozen CLIP-image encoder augmented with trainable spatial adapters(A) and temporal adapters(TA) that explicitly model inter-frame dependencies, enabling both short- and long-range temporal reasoning across the video. To guide the dual-contrast alignment, we prepend learnable context vectors ($V_i, W_i$) to class-specific text prompts---``active \{bleeding\}'', ``no active \{bleeding\}'', etc---and encoded through CLIP's frozen text encoder, producing bleeding ($t_B$) and non-bleeding ($t_N$) text embeddings. Multi-scale visual--text similarity scores $S^6, S^{12}, S^{18}, S^{24}$ are aggregated and passed through a sigmoid function to yield the final video-level bleeding prediction $\hat{y}_i$.

% The hierarchical multi-scale temporal adapters across encoder layers facilitates the fine-grained temporal reasoning required for bleeding adverse event detection.
% Spatial and temporal adapters are added at four intermediate layers of the frozen CLIP image encoder. At each extraction layer, two independent spatial and temporal adapters operate in parallel. The non-bleeding branch adapter learns to project features into a subspace aligned with non-bleeding patterns, while the bleeding branch adapter learns a complementary projection aligned with bleeding adverse events. To guide the dual-contrast alignment, we employ learnable text prompts for both branches, initialized with surgically descriptive templates---e.g., "active bleeding", "oozing bleeding" for the bleeding branch and "no bleeding", "clean of bleeding" for the non-bleeding branch---and refined during training via CoOp-style context optimization.
% Critically, the two branches do not share any parameters, they are trained to capture distinct and complementary aspects of the input, with the non-bleeding adapter specializing in recognizing the absence of bleeding adverse events and the bleeding adapter specializing in detecting their presence.

% Each spatial adapter layer produces per-frame embeddings $\mathbf{h}_{nb}^l, \mathbf{h}_{b}^l \in \mathbb{R}^{B \times T \times 768}$. These are passed through dedicated temporal adapters $\mathcal{TA}_{n}^l$ \& $\mathcal{TA}_{b}^l$ respectively.

We optimize a log-sigmoid dual-contrast loss that encourages each branch to align with its matched text prompt while repelling the opposite, summed over adapter layers $\pi \in \{6,12,18,24\}$ and both branches $c \in \{b,
n\}$: $$\mathcal{L} = -\frac{1}{B}\sum_{i,\pi,c} \log\sigma\Bigl(\ell_{i,c} \cdot \bigl(\hat{v}_{i,c}^{\pi} \cdot \hat{t}_{c} - \hat{v}_{i,c}^{\pi} \cdot \hat{t}_{\bar{c}}\bigr)\Bigr)$$
where $\hat{v}_{i,c}^{\pi}$ and $\hat{t}_c$ are L2-normalized video and text embeddings, $\bar{c}$ denotes the opposite branch, and $\ell_{i,c} \in \{-1,+1\}$ is a signed ground-truth label.

\subsubsection{HiRED: Hierarchical Entropy-Driven Key-Frame Selection} From each temporal window of $N$ frames, only $K \ll N$ key-frames are selected for the CLIP encoder. Uniform subsampling treats all frames as equally informative, often retaining redundant content (instrument occlusion, smoke, static tissue). HiRED addresses this via two-phase entropy-guided selection.
\paragraph{Phase 1: Per-Frame Scoring.}
Red channel of each frame is extracted, and the Shannon entropy~\cite{spadaro2023shannon} of the normalized 256-bin intensity histogram is computed as $S_i = -\sum_{k=0}^{255} p_i(k)\,\log_2 p_i(k)$, where $p_i(k)$ denotes the fraction of pixels with intensity $k$ in frame $i$. High entropy indicates diverse red-channel intensities—a visual cue of bleeding. This yields a score vector $\mathbf{S} \in \mathbb{R}^N$.

\begin{figure}[t]
    \centering
    \includegraphics[width=0.6\textwidth]{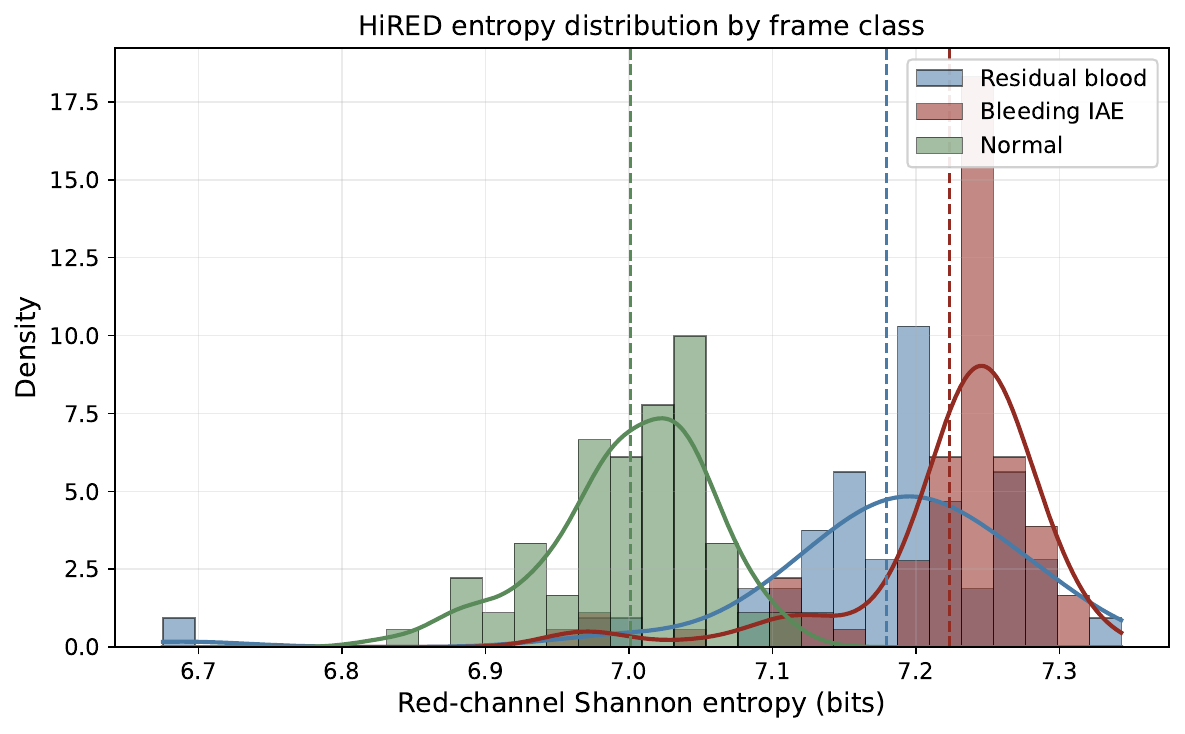}
    \caption{Red-channel Shannon Entropy Distribution.}
    \label{fig:red-channel}
\end{figure}

\paragraph{Phase 2: Hierarchical Elimination.}
Given $\mathbf{S}$, frames are iteratively pruned over multiple rounds as shown in Fig~\ref{fig:method_diagram}(b). Let $\mathcal{C}^{(t)}$ denote the candidate set at round $t$, with $\mathcal{C}^{(0)}=\{1,\dots,N\}$. 
At each round, the candidates are partitioned into $G^{(t)}$ contiguous segments of roughly equal size (segment size $s$), where
\begin{equation}
G^{(t)}=\max\!\left(K,\,\left\lfloor \frac{|\mathcal{C}^{(t)}|}{s}\right\rfloor\right)
\end{equation}

Within each segment $G_g^{(t)}$, only the highest-entropy frame($c_g^{(t)}$) survives, and thus
% : $c_g^{(t)} = \arg\max_{i \,\in\, \mathcal{G}_g^{(t)}} S_i$\allowbreak; 
$\mathcal{C}^{(t+1)} = \{c_1^{(t)}, \dots, c_{G^{(t)}}^{(t)}\}$ for the next round. The process terminates when $|\mathcal{C}^{(t)}| = K$. Early rounds eliminate low-entropy frames globally while later rounds make fine-grained distinctions within progressively smaller, higher-quality pools. Because segments are always contiguous, the final $K$ key-frames are guaranteed to span the full temporal extent of the clip. During training, a stochastic variant (HiRED-J) injects diversity by selecting the second-best frame per segment with probability $\rho{=}0.3$; at test time, selection is deterministic.

Residual-blood segments are often static and redundant, yielding lower red-channel entropy. HiRED retains high-diversity frames more indicative of active bleeding IAEs despite overlap (Fig.~\ref{fig:red-channel}), thereby suppressing residual-blood evidence within each 300-frame window and improving temporal discrimination.

\subsubsection{Multi-scale Temporal Adapter (MTA)}
To capture fine-grained temporal reasoning, we introduce MTA, a lightweight temporal adapter operating across multiple semantic scales of the frozen CLIP image encoder as shown in Fig~\ref{fig:method_diagram}. Each adapter consists of 6 self-attention layers, 12 attention heads, and learnable frame positional embeddings. Intermediate representations are extracted at four depths $\pi \in \{6,12,18,24\}$, spanning a hierarchy from low-level texture and color transitions to high-level semantic features encoding instrument-tissue interactions. At each scale, per-frame patch tokens produced by the dual spatial adapters are spatially mean-pooled into frame-level embeddings $\mathbf{F}_{i,c}^\pi = [\mathbf{f}_{i,c}^{\pi,1}, \dots, \mathbf{f}_{i,c}^{\pi,T}] \in \mathbb{R}^{T \times d}$. These are augmented with learnable temporal position embeddings and processed by a temporal transformer with a residual connection:
$$\hat{\mathbf{v}}_{i,c}^{\pi} = \mathrm{norm}\left(\mathrm{MaxPool}\Big(\mathrm{Transf}\bigl(\mathbf{F}_{i,c}^\pi + \mathbf{E}_{\mathrm{pos}}\bigr) + \mathbf{F}_{i,c}^\pi\Big)\right)$$
where $\mathbf{E}_{\mathrm{pos}} \in \mathbb{R}^{T \times d}$ are learnable temporal position embeddings, $\mathrm{Transf}(\cdot)$ is a stack of multi-head self-attention layers with feed-forward networks performing global temporal reasoning, and the residual connection preserves pretrained CLIP representations by initializing the module as a near-identity mapping. This multi-scale design allows adapter layers to learn depth-specific temporal aggregation: shallow layers capture low-level dynamics (e.g., color changes), while deeper layers model higher-level semantics like bleeding progression.

\section{Experimental Setup}
\subsubsection{Implementation}
We use CLIP ViT-L/14 architecture as backbone and 240-pixel input images. All backbone parameters are frozen during training; only the lightweight spatial adapter modules, temporal adapter modules \& learnable text prompts are optimized. Training is performed with Adam optimizer with learning rate of $1\times e^{-3}$, batch size 16 for 90 epochs. 

% We optimize a log-sigmoid loss over the dual scores $s_c^l$:
% $$\mathcal{L}=-\frac{1}{B}\sum_{i=1}^{B}\sum_{l}\sum_{c=1}^{2}\log\sigma\!\left(\ell_{i,c}\,s_{i,c}^{l}\right),\quad \ell_{i,c}\in\{+1,-1\}.$$ DEFINE COMPONENTS OF LOSS FUNCTIN HERE ??

% Describe some critical hyperparams.
\subsubsection{Baselines}
We compare against four representative state-of-the-art methods: MadCLIP~\cite{shiri2025madclip} and VadCLIP~\cite{wu2024vadclip} (dual-branch CLIP-based anomaly detection), ActionCLIP~\cite{wang2023actionclip} (temporal transformer over CLIP features), and SEDMamba~\cite{xu2024sedmamba} (a Mamba-based surgical error detection model). These baselines span both CLIP-based and domain-specific paradigms relevant to our design. BetaMixture~\cite{bose2025feature} is excluded due to the lack of publicly available code, preventing reproducible comparison.

\subsubsection{Evaluation Metrics}
We adopt standard classification metrics, including F1 score and Recall for performance evaluation. We additionally report the Matthews Correlation Coefficient (MCC), a correlation-based metric between ground-truth and predicted labels, widely regarded as a robust performance measure for imbalanced binary classification~\cite{chicco2020advantages}. MCC score ranges from $-1$ (complete disagreement) through $0$ (random prediction) to $+1$ (perfect prediction), providing a balanced measure even under class imbalance.

\begin{table}[htbp]
\centering
\caption{Performance comparison on the MultiBypass dataset and zero-shot transfer evaluation on the EndoPit-IAE dataset ($p < 0.001$).}
\label{tab:quantitative_results}
\begin{tabular}{lcccccc}
\toprule
& \multicolumn{3}{c}{\textbf{MultiBypass}} & \multicolumn{3}{c}{\textbf{EndoPit-IAE (Zero-shot)}} \\
\cmidrule(lr){2-4}\cmidrule(l){5-7}
\textbf{Method}
& \textbf{F1} $\uparrow$ & \textbf{Recall} $\uparrow$ & \textbf{MCC} $\uparrow$
& \textbf{F1} $\uparrow$ & \textbf{Recall} $\uparrow$ & \textbf{MCC} $\uparrow$ \\
\midrule
SEDMamba                  & 45.13 & 46.82 & 0.28 & 70.49 & 67.19 & 0.19 \\
VadCLIP                   & 54.31 & 69.66 & 0.38 & 63.00 & 53.91 & 0.18 \\
ActionCLIP                & 58.38 & 73.03 & 0.44 & 75.77 & \textbf{91.15} & 0.18 \\
MadCLIP                   & 58.22 & 69.66 & 0.44 & 77.00 & 88.48 & 0.27 \\
\textbf{DBT-Bleed (Ours)} & \textbf{64.91} & \textbf{78.65} & \textbf{0.53} & \textbf{83.00} & 89.53 & \textbf{0.35} \\
\bottomrule
\end{tabular}
\end{table}

\section{Results and Discussions}
Table~\ref{tab:quantitative_results} shows quantitative results on MultiBypass dataset. Our method achieves the best overall performance, surpassing the second-best ActionCLIP method by 6.53\% gain in F1 score and 5.62\% gain in Recall.
DBT-Bleed improves MCC by 9\%, indicating a clearer bleeding/non-bleeding decision boundary while accounting for all confusion-matrix entries under class imbalance.
DBT-Bleed also improves AUPRC over MadCLIP (61.6 vs. 56.29).
All improvements over the baselines are statistically significant (McNemar's exact test, $p < 0.001$ for all pairwise comparisons).

\begin{figure}[t]
    \centering
    \includegraphics[width=0.8\textwidth]{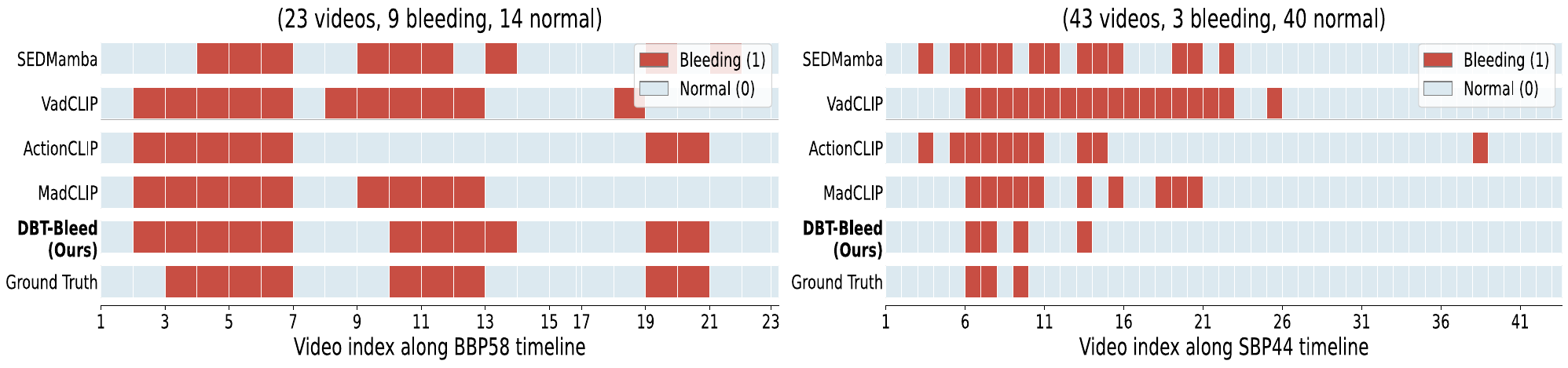}
    \caption{Ribbon visualizations of model predictions on MultiBypass dataset.}
    \label{fig:result_bar_plot}
\end{figure}

\begin{figure}[t]
    \centering
    \includegraphics[width=0.8\textwidth]{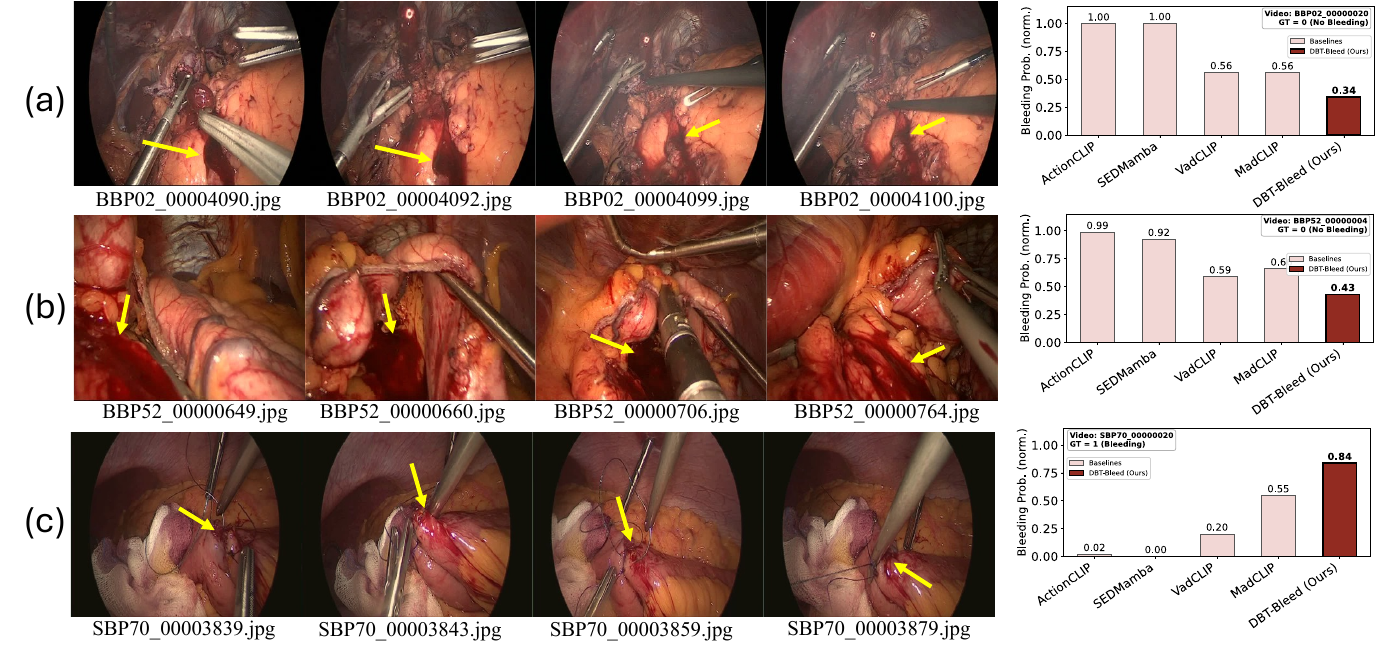}
    \caption{Qualitative MultiBypass results: (a)-(b) show non-IAE residual blood; (c) subtle bleeding IAE. Our method yields accurate, confident predictions.}
    \label{fig:Results_qualitative}
\end{figure}

We present ribbon visualizations in Fig.~\ref{fig:result_bar_plot}, enabling a direct comparison of prediction dynamics between our method and the baselines.
Fig.~\ref{fig:result_bar_plot} demonstrates that our method achieves superior bleeding IAE video-clip detection performance compared to all baselines, with improved sensitivity and specificity.
We further provide qualitative comparisons in Fig.~\ref{fig:Results_qualitative}. Notably, our method successfully identifies several critical cases with high confidence where residual blood is present on the scene, highlighting its robustness in difficult scenarios.

We further evaluate cross-procedure generalization on the in-house EndoPit-IAE dataset in a zero-shot setting. As shown in Table~\ref{tab:quantitative_results}, although EndoPit-IAE has substantial domain shift between surgery types against MultiBypass, the proposed DBT-Bleed still achieves the best zero-shot performance (6\% gain in F1 and 8\% gain in MCC score), showing the better generalization across different datasets.
% The higher zero-shot F1 and Recall score on EndoPit-IAE compared to supervised performance on MultiBypass is attributed to higher resolution of EndoPit-IAE videos with high color contrast.
The higher zero-shot F1 and Recall score on EndoPit-IAE compared to supervised performance on MultiBypass is potentially accounted by its higher resolution and stronger color contrast; importantly, under the same zero-shot protocol, DBT-Bleed consistently outperforms all baselines.

\subsubsection{Ablation Studies} We conduct experiments to select an appropriate video length. On MultiBypass, we observe that approximately 95\% of continuous adverse events span fewer than 300 frames.
Accordingly, we perform an ablation study over $N$ in $16,32,64,100,300,400,500$ frames per video. Peak performance occurs at $N=300$ as shown in Table~\ref{tab:ablation_combined}, which we adopt as the default setting.
We hypothesize that this regime best leverages the proposed HiRED key-frame selection strategy by preserving sufficient temporal context.
Furthermore, both the MTA and HiRED modules yield consistent improvements across all evaluation metrics as shown in Table~\ref{tab:ablation_combined}.

\begin{table}[t]
\centering
\caption{Ablation study. Left: video length $N$. Right: the proposed components.}
\label{tab:ablation_combined}

\resizebox{\linewidth}{!}{%
\begin{tabular}{@{}c@{\kern0.5em}c@{}}

% -------- LEFT (video lengths; transposed to reduce height) --------
\begin{tabular}{@{}lccccccc@{}}
\toprule
\textbf{$N\rightarrow$} & 16 & 32 & 64 & 100 & 300 & 400 & 500 \\
\midrule
F1     & 52.42 & 54.51 & 55.10 & 56.72 & \textbf{64.91} & 58.70 & 59.17 \\
Recall & 59.38 & 59.30 & 60.63 & 66.76 & \textbf{78.65} & 74.06 & 74.28 \\
MCC    & 0.48  & 0.50  & 0.49  & 0.50  & \textbf{0.53}  & 0.41  & 0.42  \\
\bottomrule
\end{tabular}

&

% -------- RIGHT (proposed components) --------
\begin{tabular}{@{}cccccc@{}}
\toprule
Base & MTA & HiRED & F1 & Recall & MCC \\
\midrule
\cmark & \xmark & \xmark & 58.22 & 69.66 & 0.44 \\
\cmark & \cmark & \xmark & 60.15 & 74.91 & 0.47 \\
\cmark & \cmark & \cmark & \textbf{64.91} & \textbf{78.65} & \textbf{0.53} \\
\bottomrule
\end{tabular}

\end{tabular}%
}
\end{table}

\section{Conclusions}
We present DBT-Bleed, a dual-branch multi-scale temporal framework to capture fine-grained short- and long-term bleeding progression while addressing the challenges posed by visually similar residual blood. The proposed HiRED frame selection strategy further enables efficient long-range modeling without compromising fine-grained temporal information. Our method improves F1 by 6.53\% and MCC by 9\% on the MultiBypass dataset.
We also demonstrate zero-shot cross-procedure generalization of DBT-Bleed with 6\% F1 and 8\% MCC gain on a newly curated Endonasal Pituitary Surgery dataset, EndoPit-IAE, the first neurosurgical IAE-annotated dataset. While this study focuses on binary bleeding detection, future work will develop larger, more diverse datasets and extend DBT-Bleed toward multi-label, clinically representative IAEs modeling.

\begin{credits}
\subsubsection{\ackname} This work was supported by Ministry of Education Tier 2 grant, Singapore (T2EP20224-0028), NUS-UCL Research and Innovation Collaboration Fund 2025, and an EPSRC Standard Grant [EP/Z534754/1]. 

\subsubsection{\discintname}
The authors declare no conflict of interests.
% It is now necessary to declare any competing interests or to specifically state that the authors have no competing interests. Please place the statement with a bold run-in heading in small font size beneath the (optional) acknowledgments\footnote{If EquinOCS, our proceedings submission system, is used, then the disclaimer can be provided directly in the system.}, for example: The authors have no competing interests to declare that are relevant to the content of this article. Or: Author A has received research grants from Company W. Author B has received a speaker honorarium from Company X and owns stock in Company Y. Author C is a member of committee Z.
\end{credits}

\bibliographystyle{splncs04}
\bibliography{Paper-4663}

\end{document}